\documentclass[10pt,twocolumn,letterpaper]{article}

\usepackage{cvpr}
\usepackage{times}
\usepackage{epsfig}
\usepackage{graphicx}
\usepackage{amsmath}
\usepackage{amssymb}
\usepackage{booktabs}
\usepackage{subfigure}
\usepackage{multirow}
\usepackage{algorithm}
\usepackage{algorithmic}
\newcommand{\pxj}[1]{\textcolor[rgb]{0,0,0}{#1}}

\usepackage[pagebackref=true,breaklinks=true,letterpaper=true,colorlinks,bookmarks=false]{hyperref}

\cvprfinalcopy 

\begin{document}

\title{Unsupervised Person Re-Identification with Multi-Label Learning Guided Self-Paced Clustering}

\author{Qing~Li\textsuperscript{1,3}, Xiaojiang~Peng\textsuperscript{2}, Yu~Qiao\textsuperscript{3}, Qi Hao\textsuperscript{1}\\
\textsuperscript{1}{\small School of Computer Science and Engineering, Southern University of Science and Technology, Shenzhen, China}\\
\textsuperscript{2}{\small College of Big Data and Internet, Shenzhen Technology University, Shenzhen, China} \\
\textsuperscript{3}{\small Department of Multimedia Laboratory, Shenzhen Institutes of Advanced Technology,} \\
{\small Chinese Academy of Sciences, Shenzhen, China} \\
{\small liq36@sustech.edu.cn, pengxiaojiang@sztu.edu.cn, yu.qiao@siat.ac.cn, hao.q@sustech.edu.cn}
\and
 }

\maketitle

\begin{abstract}
Although unsupervised person re-identification (Re-ID) has drawn increasing research attention recently, it remains challenging to learn discriminative features without annotations across disjoint camera views. In this paper, we address the unsupervised person Re-ID with a conceptually novel yet simple framework, termed as Multi-label Learning guided self-paced Clustering (MLC). MLC mainly learns discriminative features with three crucial modules, namely a multi-scale network, a multi-label learning module, and a self-paced clustering module. Specifically, the multi-scale network generates multi-granularity person features in both global and local views. The multi-label learning module leverages a memory feature bank and assigns each image with a multi-label vector based on the similarities between the image and feature bank. After multi-label training for several epochs, the self-paced clustering joins in training and assigns a pseudo label for each image. The benefits of our MLC come from three aspects: i) the multi-scale person features for better similarity measurement, ii) the multi-label assignment based on the whole dataset ensures that every image can be trained, and iii) the self-paced clustering removes some noisy samples for better feature learning. Extensive experiments on three popular large-scale Re-ID benchmarks demonstrate that our MLC outperforms previous state-of-the-art methods and significantly improves the performance of unsupervised person Re-ID.
\end{abstract}

\section{Introduction}
Person re-identification (Re-ID) aims at searching people across non-overlapping surveillance camera views deployed at different locations by matching person images \cite{li2014deepreid,zheng2016person,wang2016joint}. Due to its importance in smart cities and large-scale surveillance systems, person Re-ID is already a well-established research problem in computer vision \cite{fukuda2017efficient,hermans2017defense,zhong2018generalizing,kodirov2016person}. \pxj{Though great progress has been made in both benchmarks and approaches in recent years, person Re-ID remains an open challenging problem due to the difficulty of learning robust and discriminative representation with large variant intra-person appearance and high inter-person similarity.}


Over the past \pxj{decades, most of the existing person Re-ID works focus on feature designing and metric learning~\cite{liao2015person,chen2017person,xiong2014person}.}
Recently, modern deep learning has been applied to the Re-ID community and achieved significant progress \cite{li2014deepreid,cheng2016person, zheng2019pose}. 
\pxj{Most of these works tackle Re-ID in a supervised learning manner, which are limited by the small scale of Re-ID datasets. Nevertheless, collecting unlabeled pedestrian images are much easier and cheaper, thus training deep networks on large scale unlabeled data becomes increasingly necessary and practical.}

\pxj{In fact, unsupervised learning for person Re-ID has become a hot topic in more recent years~\cite{deng2018image,yu2018unsupervised,song2020unsupervised,wu2019clustering}. There mainly exist two types of unsupervised person Re-ID methods. The first one is based on unsupervised domain adaption (UDA) where the source domain is usually a labeled dataset and the target domain is an unlabeled dataset. Most of these UDA based methods use transfer learning to learn the knowledge in the labeled source Re-ID dataset and transfer them to target datasets \cite{wang2018transferable,peng2016unsupervised,yang2020part,wu2019distilled}. Specifically, some works use generative adversarial networks (GAN) for transferring sample images from the source domain to the target domain while preserving the person identity as much as possible \cite{wei2018person,zhong2019invariance,zhong2018camstyle}. Some others first train models on the source domain, then leverage self-supervised learning and clustering to 
estimate pseudo-labels on the target domain iteratively to fine-tune the pre-trained model \cite{song2020unsupervised,wu2019clustering,fan2018unsupervised,jin2020global}. 
The main disadvantages of unsupervised domain adaption Re-ID are two-fold. On the one hand, it still needs expensive labeled data and the performance is usually limited by the scale of the labeled source dataset. On the other hand, these methods ignore the sample relations between the source and the target datasets.}

\begin{figure}
    \centering
    \includegraphics[width=0.45\textwidth]{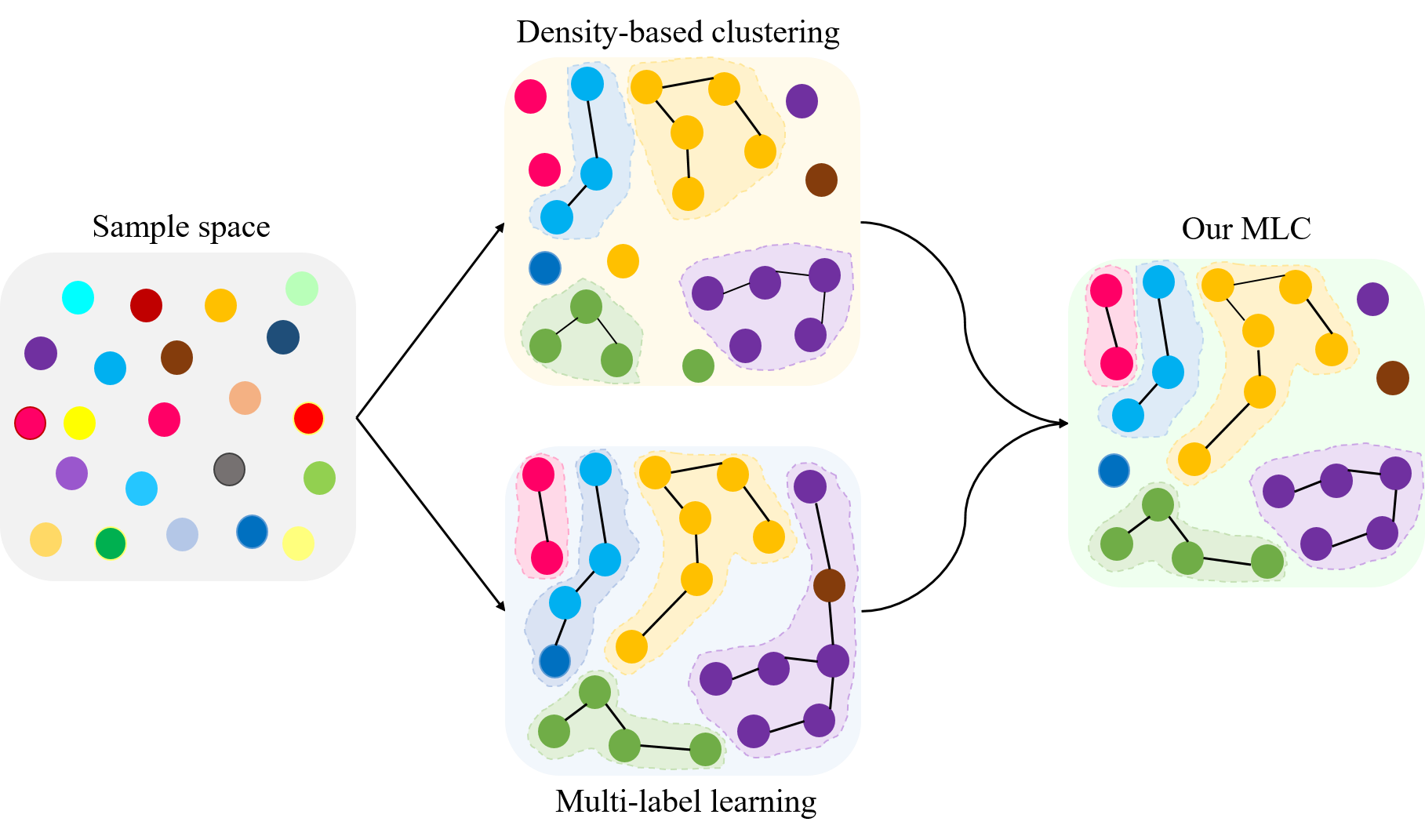}
    \caption{Comparison of recent unsupervised person Re-ID methods and our multi-label learning guided self-paced clustering (MLC) method.}
    \label{fig:moti}
\end{figure}

The second type of unsupervised Re-ID methods are based on fully unsupervised learning. Their goal is to learn discriminative representations in large scale unlabeled data. Most of these methods use clustering to generate pseudo labels.
For example, Lin \textit{et al.}~\cite{lin2019bottom} propose a bottom-up clustering (BUC) framework that trains a network with pseudo labels iteratively. The inaccurate nature of clustering algorithm on large intra-class variations makes the pseudo labels noisy, which in consequence leads to poor performance. In order to avoid wrong merging and make full use of all the images, Ding \textit{et al}.~\cite{ding2019towards} propose an elegant and practical density-based clustering approach by incorporating the cluster validity criterion. Wang \textit{et al}.~\cite{wang2020unsupervised} consider the unsupervised person Re-ID as a multi-label classification task to progressively seek true labels. They introduce a Memory-based Multi-label Classification Loss (MMCL) method, which iteratively predicts multi-labels and updates the network with multi-label classification loss. As illustrated in Figure \ref{fig:moti}, the density-based clustering strategy tries to keep high-purity samples for model training but it may ignore useful hard samples. The multi-label based strategy keeps all the samples in the memory which may introduce noisy samples in training phase.

In this paper, to address the above issues of clustering and multi-label strategies, we propose a conceptually novel yet simple framework for unsupervised person Re-ID, termed as multi-label learning guided self-paced clustering (MLC). 
Specifically, MLC learns discriminative information with three crucial modules, namely a multi-scale network (MN), a multi-label learning (ML) module, and a self-paced clustering (SC) module. 
The MN module is used to mine the multi-scale person features for better similarity measurement. Comparing to the previous methods that only extracting the global features, the MN module captures more non-salient or infrequent local information. Local feature learning is demonstrated as an effective strategy to enhance the feature representation \cite{farenzena2010person} which is complementary to the global feature \cite{wang2018learning,fu2019self}. 
The ML module generates a multi-label vector for each image based on a memory bank. Specifically, each sample in the memory bank is viewed as a single class, and a sample is assigned with a multi-hot vector where the corresponding items are activated if the sample is similar with those indexed samples in memory. To this end, these images with the same identity could get similar multi-label vectors.
To avoid training noisy samples which may hurt the final model, the SC module is added after several training epochs of ML. The SC module mainly removes noisy samples by density-based clustering algorithm and assigns pseudo labels for multi-class training. We jointly train the whole network in an end-to-end manner. 


We evaluate the proposed MLC framework on three large-scale datasets including Market-1501, DukeMTMC-reID, and MSMT17 without leveraging their annotations. Experimental results show that our MLC significantly improves the performance of unsupervised person Re-ID without any annotations and achieves performance superior or comparable to the state-of-the-art methods.
h

\section{Related work}
In this section, we review the Person Re-Identification (Re-ID) technology in the view of supervised learning, unsupervised domain adaption (UDA), and unsupervised learning.

\subsection{Supervised Learning for Person Re-ID}

Most existing person Re-ID methods employ supervised model learning on per-camera-pair manually labeled pairwise training data. The main techniques of these methods are focused on distance metric or subspace learning \cite{chen2017person,xiong2014person}, view-invariant discriminative feature learning \cite{liao2015person,wang2014unsupervised,zhao2013unsupervised}. Based on the surge of deep learning techniques, the field of supervised person Re-ID has witnessed rapid progress in recent years \cite{zhong2018generalizing,li2018harmonious,ahmed2015improved}. Generally, most of these methods assume that person images are well-aligned. In fact, it is difficult and impractical to given perfect annotation information while person poses are changing. To overcome this limitation, lots of works have adopted attentional deep learning approaches to tackle the misalignment problem. However, supervised learning methods rely on substantial costly and time-consuming labeled training data, which limits those approaches' scalability and practicability. Although these methods show certain generalization ability on labeled data, supervised Re-ID methods still lacks of effective and practical applications on unlabeled data. 

\subsection{UDA for Person Re-ID}
With the progress of deep learning on unsupervised feature learning, researchers begin to apply deep learning to address unsupervised person Re-ID tasks \cite{lin2019bottom,wu2019unsupervised}. The open set domain adaptation has been extensively applied to solve the problem of image classification tasks \cite{panareda2017open,feng2019attract}, where several classes are unknown in the two domains (or in the target domain). Recently, the domain adaptation strategy has been also widely used for unsupervised person Re-ID tasks \cite{deng2018image,peng2016unsupervised,wang2018transferable}. Most of these UDA methods refer to transfer learning to learn the knowledge from the source dataset and transfer them to target dataset. It is clear that the former is an auxiliary and necessarily labelled dataset but the latter is a unlabelled dataset \cite{yu2018unsupervised,wu2019distilled,kodirov2015dictionary,yu2019unsupervised}. However, classes of the two domains are entirely different for UDA in person Re-ID, which presents a greater challenge.

To address this domain adaptation problem, there are three typical methods as follows. The first category of methods explore image-style transformation from labeled source domain to unlabeled target domain \cite{deng2018image,wang2018transferable,wei2018person}. In \cite{wei2018person}, PTGAN enforces the self-similarity of an image before and after translation and the domain dissimilarity of a translated source image and a target image. In paper \cite{zhong2018generalizing}, Zhong \textit{et al.} first propose a Hetero-Homogeneous Learning (HHL) method to learn camera-invariant network for the target domain. However, HHL overlooks the latent positive pairs in the target domain, which might lead the Re-ID model to be sensitive to the background or pose variations in the target domain. To overcome these drawbacks, Zhang \textit{et al.} \cite{zhang2019self} propose a self-training method with a progressive augmentation framework to promote the model performance progressively on the target dataset.

The second category of methods utilize a model distillation by using the teacher model to guide learning of the student model. A vast majority of knowledge distillation methods are adopted for Re-ID problems to alleviate cross-camera scene variation explicitly or implicitly \cite{fukuda2017efficient,wu2019distilled}. In \cite{wu2019distilled}, a multi-teacher adaptive similarity distillation framework is proposed to learn a user-specified lightweight student model from multiple teacher models, without access to source domain data. Wu et al. \cite{wu2019unsupervised} propose to learn consistent pairwise similarity distributions for intra-camera and cross-camera matching with the guidance of prior common knowledge of intra-camera matching.

The third category of methods attempts on optimizing the Re-ID model with soft labels for target-domain samples by measuring the similarities with reference images or features. Zhong \textit{et al.} \cite{zhong2019invariance} investigate the impact of intra-domain variations and impose three types of invariance constraints on target samples. They assign soft labels and minimize the target invariance by proposing an exemplar memory module \cite{he2020momentum,wang2020cross}, which caches feature vectors of every instance. And then, Yu \textit{et al.} \cite{yu2019unsupervised} propose a method called deep soft multilabel reference learning (MAR) method, which conducts multiple soft-label learning by comparing with a set of reference persons. Ge \textit{et al.} \cite{ge2019mutual} designed an asymmetrical framework to generate more robust soft labels via the mutual mean-teaching. However, these methods need to learn one or several effective teacher model(s), which only depend(s) on the diversity and quality of source domain data. It can't be ignored that they still need a labeled source data, and they don't explore the sample similarity between the source and the target domain. However, our method not only discard the requirment of labeled source domain data, but also could mine sample similarities on target domain data directly. Specially, we leverage MN to mine multi-granularity person features and ML module to store these up-to-date multi-scale features for the whole target data.

\begin{figure*}[tp]
\centering
\includegraphics[width=0.9\textwidth]{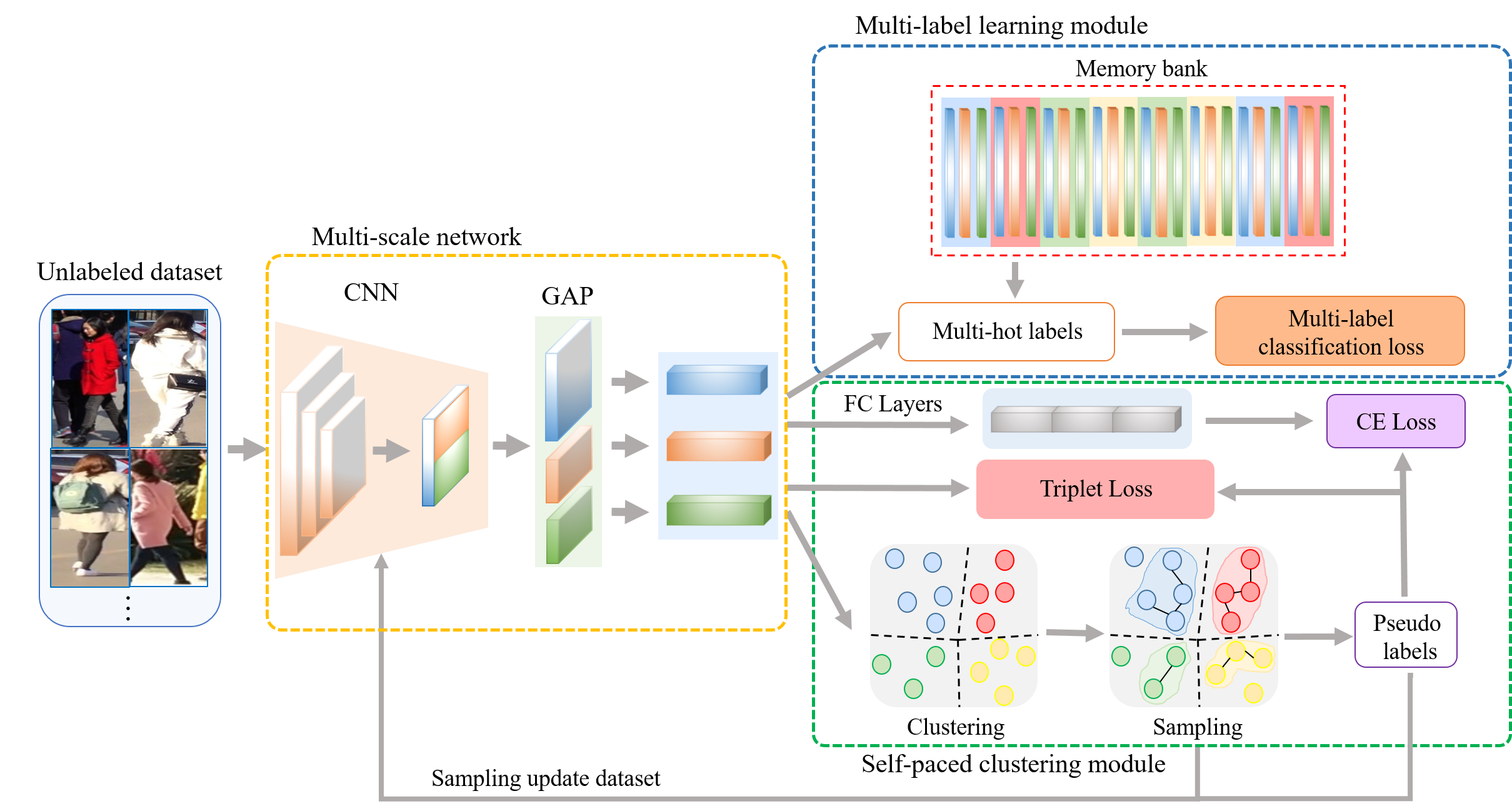}
\caption{\pxj{The pipeline of our MLC framework for unsupervised person Re-ID. It mainly contains three sub-components ( indicated by three colored dot rectangles): 1) multi-scale network (MN), 2) multi-label learning (ML) module, and 3) self-paced clustering (SC) module. Given an image, the multi-scale network extracts both global feature in the whole image and local features in sub regions by global max pooling. The ML module leverages an updatable memory bank to assign multi-label vectors for all unlabeled images and uses multi-label classification loss for training. The SC module filters noisy samples and assign pseudo labels by the step-wise clustering and sampling.}
}
\label{fig:framework}
\end{figure*}

\subsection{Unsupervised Learning for Person Re-ID.}

Unsupervised learning methods have attracted much attention because of their capability of saving the cost of manual annotations. The main task of unsupervised learning has shifted to how to fully mine the useful information in the increasingly growing unlabeled dataset. Unsupervised learning methods for person Re-ID generally involve two aspects: traditional unsupervised methods and clustering-guided deep learning methods.

Traditional unsupervised person Re-ID mainly focus on feature learning, which created hand-craft features \cite{bazzani2013symmetry} that can be utilized directly to the unlabeled dataset. In the earlier works, Liao \textit{et al.} \cite{liao2015person} propose the local maximal occurrence (LOMO) descriptor, which includes the color and SILTP histograms. Zheng \textit{et al.} \cite{zheng2015scalable} propose to extract the 11-dim color names descriptor for each local patch, and aggregate them into a global vector through a Bag-of-Words model. It is worth pointing out that these methods' performance are not effective and satisfactory on unlabelled dataset. The latest studies suggest that it is difficult to learn robust and discriminative features by traditional unsupervised methods.

With the surge of deep learning techniques, recent studies have focused on the clustering-guided deep learning methods for unsupervised Person Re-ID \cite{yu2018unsupervised,yu2017cross,wang2016towards}. Because of the most important is that clustering method is intuitive and efficient for unsupervised machine learning. These methods generally focused on generates pseudo-labels on the target domain and then use these pseudo-labels to learn deep models in a supervised manner. 
Liu \textit{et al.} \cite{liu2017stepwise} propose a stepwise metric promotion approach to refine the pseudo labels by iteratively estimate the annotations of training tracklets. Wu \textit{et al.} \cite{wu2018exploit} propose a progressive sampling method to gradually predict reliable pseudo labels and uncover the unlabeled data for one-shot video-based Re-ID. Yang \textit{et al.} \cite{yang2020asymmetric} introduced the asymmetric co-teaching strategy to refine the pseudo-labels in the clustering-based method. Zhai \textit{et al.} \cite{zhai2020ad} present a novel augmented discriminative clustering technique that incorporates style-translated images to improve the discriminativeness of instance features. However, such methods rely on a good deep Re-ID model as an initialized feature extractor for unsupervised learning. Aside from these methods that require an auxiliary 
Re-ID model, they also face the hard or difficult samples to effect the quality of the label predict. Therefore, some researchers first consider focusing on the fully unsupervised Re-ID task without rely on any initialized model.

Different from previous works, Lin \textit{et al.} \cite{lin2019bottom} proposed a bottom-up clustering framework that iteratively trains a network based on the pseudo labels generated by unsupervised clustering. It not only considers the diversity over each sample but also exploits the similarity within each class. Ding \textit{et al.} \cite{ding2019towards} designs a novel dispersion-based clustering approach which can discover the underlying feature space for unlabeled pedestrian image data. Zeng \textit{et al.} \cite{zeng2020hierarchical} propose a hierarchical clustering with hard-batch triplet loss. These approaches have explored cluster distributions in the target domain. They still face the challenge on how to precisely predict the label of hard samples. On Different from these methods which classify each image into a single class, the multi-label classification has the potential to exhibit better efficiency and accuracy. For example, Lin \textit{et al.} \cite{lin2020unsupervised} proposed an unsupervised Re-ID network that softened labels to reflect the image similarity and eliminated the hard quantization error. Wang \textit{et al.} \cite{wang2020unsupervised}  consider the unsupervised person ReID as a multi-label classification task to iteratively predict multi-class labels and update the network with multi-label classification loss. However, this strategy of multi-label keeps all the samples in the memory which may introduce noisy samples in training phase. Different from all the above methods, our method addresses the fully unsupervised person Re-ID problem with joint multi-label and self-paced clustering. 

\section{Methodology}
In this section, we first provided an overview for our method, and then present the preliminary of unsupervised person Re-ID. Finally, we present the individual modules of our framework and the training strategy.
\subsection{Overview}
\pxj{To tackle the unsupervised person Re-ID, we propose a multi-label learning guided self-paced clustering (MLC) framework as shown in Figure \ref{fig:framework}. Our MLC framework includes three crucial modules, namely a multi-scale network (MN), a multi-label learning (ML) module, and a self-paced clustering (SC) module.
Given an image, the multi-scale network extracts both global features in the whole image and local features in sub regions by global max pooling. The ML module leverages an updatable memory bank to assign multi-hot labels for all unlabeled images and uses multi-label classification loss for training.
Commonly, the memory bank is composed of the multi-scale features of all samples in a dataset. The multi-hot label for an image is determined by the similarities between the image feature and memory bank.
The SC module filters noisy samples and assign pseudo labels by the step-wise clustering and sampling. 
In practice, we first perform several epochs of multi-label training to ensure every image is used for training and then we jointly apply SC and ML which may trade-off the noisy samples and hard samples.}

\subsection{Preliminary and Initialization}

\textbf{Preliminary.}
In fully unsupervised person Re-ID tasks, we only have an unlabeled training dataset $X = \left \{x_{1}, x_{2},\cdots,x_{N}\right \}$ containing $N$ person images. Our purpose is to learn a discriminative feature extractor $\phi(\theta; x_{i})$ from $X$ without any available annotations. The parameters of $\phi$ are optimized iteratively using an objective function. This feature extractor can be applied to the gallery set, $G=\left\{g_{1}, g_{2},\cdots,g_{N_{t}}\right\}$ of $N_{t}$ images, and the query set $Q=\left\{q_{1}, q_{2},\cdots,q_{N_{q}}\right\}$ of $N_{q}$ images. During the evaluation, for any query person image $q$, the feature extractor is expected to produce a feature vector to retrieve image $g$ containing the same person from a gallery set $G$. In other words, we need to use the features of a query image $\phi(\theta; q_{i})$ to search more similar feature with $g$ from the gallery set $G$. Hence, it is critical to learn a disciminative feature extractor $\phi(\theta;\cdot)$ for person Re-ID model. The conceptual optimization goal of distance between each pair of images is defined as,

\begin{equation}\label{eq:dist}
\hat{g}=arg\min_{g\in G}dist(\phi(\theta; q_{i}),\phi(\theta; g_{i}))
\end{equation}
where $dist(\cdot)$ is the distance metric, e.g., the L2 distance. Generally, person Re-ID mainly adopts Euclidean or cosine distances as the re-ranking method for the retrieval stage. In this case, we use the Euclidean as the distance metric function to re-rank the ReID results. More important, we will bring in a k-reciprocal encoding method with the Jaccard distance of probe and gallery images to computer the distance metric as the re-ranking method \cite{zhong2017re} for our self-paced clustering. The detailed application will be introduced in Section \ref{sec:SCM}.

\textbf{Initialization with hard labels.}
In order to learn a discriminative feature extractor, the transitional supervised learning method needs person identity labels for each image. However, there are no manually annotated labels in fully unsupervised Re-ID tasks, so we need to generate the pseudo labels instead of that. Thus, we start by treating each training image $x_{i}$ as an individual class, and initially assign $ x_{i}$ with a label $y_{i}$ by its index, i.e, $Y=\left \{y_{1}, y_{2},\cdots,y_{N}\right \}$. The feature extractor $\phi(\theta;\cdot)$ is appended by a classifier $f(w;\phi)\in \mathbb{R}^{N}$ parameterized by $w$. The optimization is defined by the following objective function:

\begin{equation}\label{eq:ce1}
\min_{\theta,w}\sum_{i=1}^{N}\l(f(w;\phi(\theta;x_{i})),y_{i}),
\end{equation}
where $l$ is the cross-entropy (CE) loss for classification. 
The hard labels Y are suitable for initialization but not reliable for unsupervised feature learning.
We warm up the neural network with the initialized hard label, allowing the model to reach a certain local optimal field where subsequential approaches need to be explored. 

\subsection{Multi-scale Network}
\begin{figure}[t]
\centering
\includegraphics[width=0.45\textwidth]{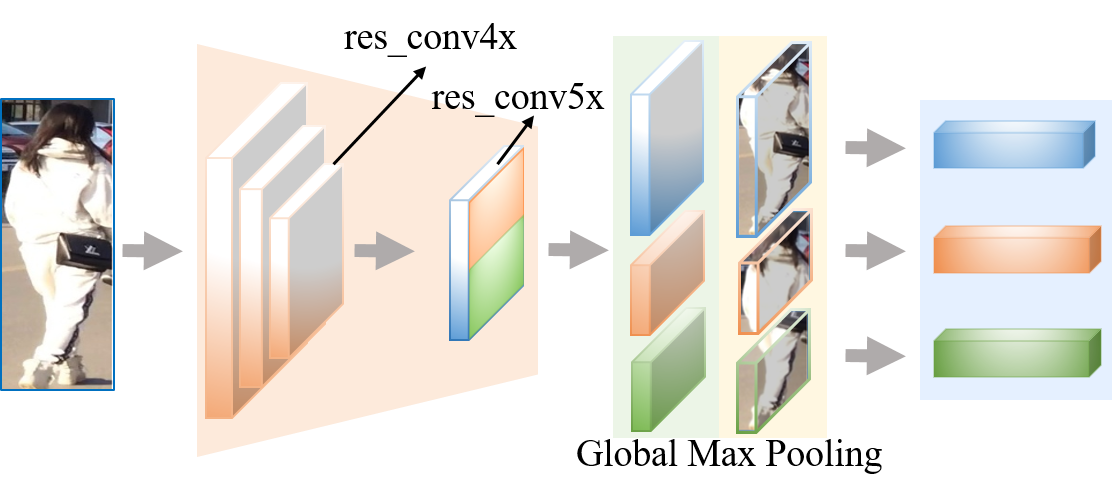}
\caption{The architecture of our multi-scale network.}
\label{fig:MN}
\end{figure}

\pxj{As the feature extractor, the multi-scale network aims to capture multi-granularity person features for similarity computing. Specifically, we use the base of ResNet50 as our backbone since it is widely adopted in person Re-ID tasks and obtains better performance. 
Inspired by the recent domain adapted Re-ID work \cite{fu2019self}, we compute the similarity between two persons not only by global information from the whole body but also by local information from different parts of a person. The detailed architecture of our MN is illustrated in Figure \ref{fig:MN}.}

\pxj{We remove the down-sampling operations before $res\_conv5\_1$, and uniformly split the feature maps into an upper and an bottom part at the last $conv$ layer. We utilize  global average pooling (GAP) operation on the whole final feature map and the partial feature maps to obtain three feature vectors for each image, \textit{i.e.}, $f_g, f_{up}, \textrm{and} ~f_{low}$. To obtain better multi-granularity features, we concatenate them as a final person representation $f_{all}$.}

\subsection{Multi-label learning module}
\pxj{The multi-label learning (ML) module aims to learn discriminative features by assigning similar multi-hot labels for similar images based on a feature memory bank. }

\textbf{Memory Bank}.
A memory bank consists of the representations of all samples in the dataset. Following \cite{wang2020unsupervised,he2020momentum,wang2020cross}, we maintain the memory bank to serve as a feature storage that saves up-to-date features of the training dataset. Memory bank allows the network discovering more negative samples from the memory buffer to pair with positive sample without recomputing their features Compared with previous methods, the benefit of a memory bank is to collect more feature informative pairs with the cost of memory space for features stored in the memory bank.

\textbf{Initialization}.
We initialize the memory module $\boldsymbol{M}$ by computing the features of a set of randomly sampled training images based on the warm-up model. Formally, $\boldsymbol{M}=\left \{ (f^{1}_{all}),(f^{2}_{all}),\cdots,(f^{N}_{all}) \right \}$, where the $\boldsymbol{M} \in \mathbb{R}^{N\times d}$ and $d$ is the dimension of these features. The $f^{i}_{all}$ is initialized as the feature of the $i$-$th$ sample $x_{i}$. In the memory bank, each slot $\boldsymbol{M\left[i\right]}$ stores the L2-normalized feature $f^{i}_{all}$ in the key part, while storing the hard label in the value part.

\textbf{Updating}.
During each training iteration, the feature vectors on each mini-batch would be involved in memory bank updating. The whole unlabeled data can be cached in the memory bank, which is dynamically updated with features computed in the training. We update the memory bank in a running-average manner as follows,

\begin{equation}\label{eq:Memory}
\begin{array}{l}
\boldsymbol{M\left[i\right]^{t}}\leftarrow \alpha \boldsymbol{M\left[i\right]^{t}} +(1-\alpha )f^{i}_{all}, \\ \boldsymbol{M\left[i\right]^{t}} = \boldsymbol{M\left[i\right]^{t}} / || \boldsymbol{M\left[i\right]^{t}} ||_{2},
\end{array}
\end{equation}
where the superscript $t$ is denoted as the current iteration epoch, $\alpha$ is the momentum of updating rate. Then, we utilize the Memory-based Positive Label Prediction (MPLP) method \cite{wang2020unsupervised} with the hard label to predict the multi-hot label $\bar{y}$ based on the memory bank $\boldsymbol{M}$. Followed the MPLP, we first computes a rank list $R_{i}$ to store the similarities between a sample and the memory bank as follows,
\begin{equation}\label{eq:Rank}
R_{i}= \mathop{\textup{arg}\,\textup{sort}}_{j}(s_{i,j}), j\in [1,N],
\end{equation}
\begin{equation}\label{eq:Rank2}
s_{i,j}=\boldsymbol{M\left[i\right]^{T}} \times \boldsymbol{M\left[j\right]}
\end{equation}
where $s_{i,j}$ is the similarity score of $x_{i}$ and $x_{j}$, and the $R_{i}$ aims to find the candidates for reliable labels about $x_{i}$. We use a similarity threshold and a cycle consistency scheme to select relevant label candidates and filter hard negative labels. The positive label is set as,

\begin{equation}\label{eq:positive}
P_{i}=\hat{R}_{i}[1:l]
\end{equation}
where $\hat{R}_{i}$ is the $top$-$k_{i}$ nearest labels and $\hat{R}_{i}={R}_{i}[1:k_{i}]$, ${R}_{i[k_{i}]}$ is the last label with similarity score higher than a threshold which decide the quantity of label candidates. The $l$ satisfies $i\in R_{\hat{R}_{l}} \& i\notin R_{\hat{R}_{l+1}}$. As $P_{i}$ contains $l$ labels, $x_{i}$ would be assigned with a multi-label $\bar{y_{i}}$ in which value 1 indicate positive classes,

\begin{equation}\label{eq:multi-label}
\bar{y_{i}}=\left\{\begin{array}{l}
 \ \ 1 \quad j\in P_{i}\\
 -1  \quad j\notin P_{i}
\end{array}\right.
\end{equation}

The multi-label classification loss, which is computed on positive classes and sampled hard negative classes, is shown below.

\begin{equation}\label{eq:multi-label_loss}
\mathcal{L}_{mmcl}=\sum_{i=1}^{N}\frac{\delta }{\left | P_{i} \right |}\sum_{p\in P_{i}}\mathit{l}(p|x_{i}))+\frac{1 }{\left | S_{i} \right |}\sum_{s\in S_{i}}\mathit{l}(s|x_{i}))
\end{equation}
where $S_{i}$ is the collection of hard negative classes for $x_{i}$, we also select the $top$-$r\%$ classes as the hard negative classes, and $\left | S_{i} \right |= (N-\left | P_{i} \right |) \cdot r\%$. The $\delta$ is a a coefficient measuring the importance of multi-label classification loss.

\subsection{Self-paced clustering module} \label{sec:SCM}
\pxj{As mentioned above, the multi-label learning strategy keeps all samples for training which may hurt feature learning due to noisy samples. The self-paced clustering (SC) module is proposed to update the training dataset with more clean samples and assign pseudo labels for training. 
}

As shown in the bottom-right of Figure \ref{fig:framework}, we firstly employ the SC module to cluster the final similarity features extracted by our MN. The clustering method aims to group similar entities together after computing the distance metric with k-reciprocal encoding \cite{zhong2017re} for all training samples. The sampling is used to filter noisy samples and assigns pseudo-labels $\tilde{y}_{i}$ according to the group entities. Finally, we form a new training dataset with pseudo-labels to leverage SC and ML module for jointly learning.

For clustering methods, the selection of nearest neighbors is crucial. It is used for merging all the instances into right cluster and finally affects the clustering results and the quality of pseudo labels. The conventional k-means clustering algorithm comes as a natural choice which selects nearest neighbors according to the distance of a sample to the centroids of clusters. As an alternative choice, the DBSCAN \cite{ester1996density} algorithm can be used which selects nearest neighbors by Jaccard distance \cite{zhong2017re} matrix and  k-reciprocal nearest neighbor method. We use DBSCAN as our default SC module, and apply step-wise clustering and sampling to remove the noisy samples of the whole dataset. During the progressive clustering stage, we try to keep the most reliable clusters. 
In order to avoid accumulating training error caused by noisy clusters, we remove these noisy samples and constitute a new training set with pseudo labels for joint training.

Since labels are available after self-paced clustering, we treat the training process as a classification problem. Specifically, we adopt Cross-Entropy (CE) loss for classification and triplet loss for metric learning. We apply CE loss with label smoothing as follows,

\begin{equation}\label{eq:ce2}
\mathcal{L}_{CE_{p}}= \frac{1}{N}\sum_{i=1}^{N}\l(f(w;\phi(\theta;x_{i})),\tilde{y}_{i}^{s})
\end{equation}
where the smooth j-th item $\tilde{y}_{j,i}^{s}=1-\varepsilon+\frac{\varepsilon}{C}$ if $j=\tilde{y}_{i}$, otherwise $\tilde{y}_{j,i}^{s}=\frac{\varepsilon}{C}$. $C$ is the numbers of predict the identities by SC module. $\varepsilon$ is a small constant for the label smoothing.

Formally, the triplet loss function is defined as follows,

\begin{equation}\label{eq:triple}
\begin{split}
\mathcal{L}_{tri}= \frac{1}{N}\sum_{i=1}^{N}\max(0,\left \| f(w;\phi(\theta;x_{i}))-f(w;\phi(\theta;x_{i,p})) \right \| \\
+m-\left \| f(w;\phi(\theta;x_{i}))-f(w;\phi(\theta;x_{i,n})) \right \|)
\end{split}
\end{equation}
where $\left \| \cdot  \right \|$ denotes the $L_{2}$-norm distance, the $x_{i,p}$ and $x_{i,n}$ indicate the hardest positive and hardest negative feature sample in each mini-batch, and $m$ denotes the triplet distance margin, and sets the default value to 0. The overall loss function for optimization is the combination of multi-label classification loss and the pseudo label classification loss (CE loss and triplet loss) as follows,

\begin{equation}\label{eq:triple}
\mathcal{L}_{o}=\lambda_{1} \mathcal{L}_{mmcl}+ \lambda_{2}  (\mathcal{L}_{CE_{p}}+\mathcal{L}_{tri}).
\end{equation}

\section{Experiments}
\subsection{Datasets and settings}
We evaluate our approach on three large-scale bench-mark datasets: Market1501 \cite{zheng2015scalable}, DukeMTMC-reID \cite{zheng2017unlabeled} and MSMT17\cite{wei2018person}.

\textbf{Market1501 Datasets} Market1501 contains 1,501 person identities with 32,668 images which are captured by six cameras. It contains 12,936 images of 751 identities for training and 19,732 images of 750 identities for testing.

\textbf{DukeMTMC-reID Datasets} DukeMTMC-reID is a subset of the
DukeMTMC dataset. It has 1,404 person identities from eight cameras, with 36411 labeled images. It contains 16,522 images of 702 identities for training, and the remaining images out of another 702 identities for testing, including 2,228 images for query, and 17,661 images for gallery.

\begin{table*}[htp]
\centering
\caption{Performance ($\%$) comparison of our framework with the baseline methods on Market-1501 dataset and DukeMTMC-reID.}\label{tab:table1}
\footnotesize
\begin{tabular}{c|c|c|c|c|c|c|c|c|c|c}
\toprule
 \hline
 \multicolumn{1}{c|}{\multirow{2}*{Method}} & \multicolumn{5}{c|}{ Market-1501} & \multicolumn{5}{c}{DukeMTMC-reID} \\
 \cline{2-11}
  \multicolumn{1}{c|}{} & source & Rank-1 & Rank-5 & Rank-10 & mAP & source & Rank-1 & Rank-5 & Rank-10 & mAP \\
 \hline
 Baseline1: fully-supervised \cite{liao2015person} & Supervised & 88.5 & 96.5 & 97.9 & 70.7& Supervised & 74.8 & 87.1 & 91.5 & 58.0 \\
 Baseline2: ImageNet model & None & 8.1 & 17.5 & 23.6 & 2.2 & None & 5.6 & 11.5 & 14.9 & 1.6 \\
 Baseline3: MMLC \cite{wang2020unsupervised} & None & 79.8 & 88.4 & 91.6 & 44.7 & None & 65.6 & 75.9 & 80.1 & 39.6  \\
 \hline
 MLC w/o MN & None & 85.2 & 92.2 & 94.6 & 62.6 & None & 71.9 & 81.2 & 84.4 & 50.0  \\
 MLC & None & \textbf{86.7} & \textbf{93.5} & \textbf{95.6} & \textbf{66.2} & None & \textbf{73.6} & \textbf{82.3} & \textbf{85.5} & \textbf{52.3} \\
 \hline
\bottomrule
\end{tabular}
\end{table*}

\textbf{MSMT17 Datasets} MSMT17 is the most challenging and the currently largest-scale dataset which contains 126,441 images of 4,101 person identities captured from 15 camera views. It is spitted to 32,621 images of 1,041 identities for training, and 93,820 images of 3,060 identities for testing.

\textbf{Evaluation Metrics}
We followed the standard training/test split and evaluated the single-query test evaluation settings. For the evaluation metrics, we used the Rank-1/Rank-5 matching accuracy, which means the query picture has the match in the top-k list. And we use the mean Average Precision (mAP), which is computed from the Cumulated Matching Characteristics (CMC) \cite{gray2007evaluating}.

\textbf{Implementation Details}
We implement our method with Pytorch. For data pre-processing, input images are resized into $256 \times 128$. We apply some commonly-used data augmentation methods including random horizontal flipping, random cropping, color jitter, random erasing, and CamStyle \cite{zhong2018camera}. For a fair comparison, we adopt ResNet-50 pre-trained on ImageNet as our backbone network. The batch size is set to 128 for training. We use the standard SGD as an optimizer, the momentum is 0.9, and weight decay is $5 \times 10^{-4}$. The initial learning rate is 0.1. We train the model for $60$ epochs, and the learning rate divided by 10 after every $30$ epochs. Following \cite{wang2020unsupervised}, we initialize the memory bank as all zeros, and use the hard label to warm-up model and the memory is fully updated for $5$ epochs. The memory updating rate $\alpha$ starts from 0 and grows linearly to 0.5. The similarity threshold in ML module is 0.6. We jointly train the ML and SC modules after $15th$ epoch.

\subsection{Comparison with the baseline}
To investigate the effect of the joint learning based on our proposed MLC, we compare MLC with three baselines: 1) the first is basic supervised learning method \cite{liao2015person} with triplet loss on the labeled data; 2) the second is direct feature evaluation with the pre-trained ResNet-50 model on ImageNet; 3) the third is unsupervised learning with multi-label classification loss \cite{wang2020unsupervised}. The experimental results are shown in Table \ref{tab:table1}. The first two baselines represent the upper and lower bound performance of the backbone model. The third baseline utilizes similar multi-label prediction with ours for discriminative feature learning. In theory, when our generated pseudo labels are closer to the truth labels, the performance of our MLC will approach to the fully-supervised baseline1. 
For fair comparisons, we also present the results obtained without MN module, \textit{i.e.}, the same feature extractor with baseline methods. 

As can be seen, the baseline1 achieves the highest performance with supervised learning, $e.g.$, 70.7\% mAP on Market-1501 and 58.0\% mAP on DukeMTMC-reID. On the contrary, the performance of baseline2 is very poor which shows that there is a large gap between person Re-ID and the object classification of ImageNet. Based on the predicted multi-label, the baseline3 boosts the baseline2 significantly, and achieves 42.5\% and 38\% improvements in mAP on Market-1501 and DukeMTMC-reID, respectively. Without the multi-scale network, our MLC already outperforms all baseline methods largely, with 62.6\% mAP on Market-1501 and 50.0\% mAP on DukeMTMC-reID. By considering the global and local information (\textit{i.e.}, with MN), the MN module further boosts performance by 3.6\% mAP on Market-1501 and 2.3\% mAP on DukeMTMC-reID.

\begin{table*}[htp]
\centering
\caption{Performance ($\%$) comparison of our framework with state-of-the-art methods on Market-1501 dataset and DukeMTMC-reID.}\label{tab:table2}
\begin{tabular}{l|c|c|c|c|c|c|c|c|c|c}
\toprule
 \hline
 \multicolumn{1}{c|}{\multirow{2}*{Method}} & \multicolumn{5}{c|}{ Market-1501} & \multicolumn{5}{c}{DukeMTMC-reID} \\
 \cline{2-11}
  \multicolumn{1}{c|}{} & source & Rank-1 & Rank-5 & Rank-10 & mAP & source & Rank-1 & Rank-5 & Rank-10 & mAP \\
 \hline
 LOMO \cite{liao2015person} & None & 27.2 & 41.6 & 49.1 & 8.0 & None & 12.3 & 21.3 & 26.6 & 4.8 \\
 BoW \cite{zheng2015scalable} & None & 35.8 & 52.4 & 60.3 & 14.8 & None & 17.1 & 28.8 & 34.9 & 8.3 \\
 \hline
 BUC \cite{lin2019bottom} & None & 66.2 & 79.6 & 84.5 & 38.3 & None & 47.4 & 62.6 & 68.4 & 27.5  \\
 SSL \cite{lin2020unsupervised} & None & 71.7 & 83.8 & 87.4 & 37.8 & None & 52.5 & 63.5 & 68.9 & 28.6  \\
 DBC \cite{ding2019towards} & None & 69.2 & 83.0 & 87.8 & 41.3 & None & 51.5 & 64.6 & 70.1 & 30.0  \\
 HCT \cite{zeng2020hierarchical} & None & 80.0 & 91.6 & 95.2 & 56.4 & None & 69.6 & 83.4 & 87.4 & 50.7  \\
 MMCL \cite{wang2020unsupervised} & None & 80.3 & 89.4 & 92.3 & 45.5 & None & 65.2 & 75.9 & 80.0 & 40.2  \\
 \hline
 MLC w/o MN & None & 85.2 & 92.2 & 94.6 & 62.6 & None & 71.9 & 81.2 & 84.4 & 50.0  \\
 MLC & None & \textbf{86.7} & \textbf{93.5} & \textbf{95.6} & \textbf{66.2} & None & \textbf{73.6} & \textbf{82.3} & \textbf{85.5} & \textbf{52.3} \\
 \hline \midrule
 UMDL \cite{peng2016unsupervised} &Duke & 34.5 & 52.6 & 59.6 & 12.4 & Market & 18.5 & 31.4 & 37.4 &7.3  \\
 CAMEL \cite{yu2017cross} & Duke & 54.5 & 73.1 & - & 26.3 & Market & 40.3 & 57.6 & - & 19.8\\
 PUL \cite{fan2018unsupervised} & Duke & 45.5 & 60.7 & 66.7 & 20.5 & Market & 30.0 & 43.4 & 48.5 & 16.4 \\
 PTGAN \cite{wei2018person} & Duke & 38.6 & 57.3 & 66.1 & 15.7& Market & 27.4 & 43.6 & 50.7 & 13.5  \\
 SPGAN+LMP \cite{deng2018image} & Duke & 57.7 & 75.8 & 82.4 & 26.7 & Market & 46.4 & 62.3 & 68.0 & 26.2 \\
 MMFA \cite{lin2018multi}& Duke & 56.7 & 75.0 & 81.8 & 27.4 & Market & 45.3 & 59.8 & 66.3 & 24.7 \\
 TJ-AIDL \cite{wang2018transferable} & Duke & 58.2 & 74.8 & 81.1 & 26.5 & Market & 44.3 & 59.6 & 65.0 & 23.0 \\
 HHL \cite{zhong2018generalizing} & Duke & 62.2 & 78.8 & 84.0 & 31.4 & Market & 46.9 & 61.0 & 66.7 & 27.2 \\
 ECN \cite{zhong2019invariance} & Duke & 75.1 & 87.6 & 91.6 & 43.0 & Market & 63.3 & 75.8 & 80.4 & 40.4 \\
 MAR \cite{yu2019unsupervised}& MSMT & 67.7  & 81.9 & -  & 40.0 & MSMT & 67.1 & 79.8 & - & 48.0 \\
 PAUL \cite{yang2019patch} & MSMT & 68.5  & 82.4 & 87.4 & 40.1 & MSMT & 72.0 & 82.7 & 86.0 & 53.2 \\
 SSG \cite{fu2019self} & Duke & 80.0 & 90.0 & 92.4  & 58.3 & Market & 73.0 & 80.6 & 83.2  & 53.4 \\
 CR-GAN \cite{chen2019instance} & Duke & 77.7 & 89.7 & 92.7 & 54.0 & Market & 68.9 & 80.2 & 84.7 & 48.6 \\
 CASCL \cite{wu2019unsupervised} & MSMT & 65.4 & 80.6 & 86.2  & 35.5  & MSMT & 59.3 & 73.2 & 77.5 & 37.8 \\
 PDA-Net \cite{li2019cross} & Duke & 75.2 & 86.3 & 90.2  & 47.6 & Market & 63.2 & 77.0 & 82.5 & 45.1  \\
 MMCL \cite{wang2020unsupervised} & Duke & 84.4 & 92.8 & 95.0 & 60.4 & Market & 72.4 & 82.9 & 85.0 & 51.4  \\
 ADTC \cite{wuattention2020} & Duke & 79.3 & 90.8 & 94.1 & 59.7 & Market & 71.9 & 84.1 & 87.5 & 52.5 \\
 D-MMD \cite{mekhazni2020unsupervised} & Duke & 70.6 & 87.0 & 91.5 & 48.8 & Market & 63.5  & 78.8  & 83.9  &  46.0  \\

\hline
MLC w/o MN & Duke & \textbf{88.4} & \textbf{94.6} & \textbf{96.2} & 64.7 & Market & 73.0 & 82.9 & 85.9 & 53.8  \\
MLC & Duke & 85.6 & 93.9 & 96.0 & \textbf{65.9} & Market &\textbf{74.1} & \textbf{83.8} & \textbf{86.3} & \textbf{55.0}  \\ 

 \hline
\bottomrule
\end{tabular}
\end{table*}

\subsection{Comparison with the state-of-the-art methods}
We compare our approach with the state-of-the-art unsupervised learning methods for person Re-ID including: 1) the hand-crafted features (including LOMO \cite{liao2015person} and BoW \cite{zheng2015scalable}; 2) the pseudo label learning methods without any other labeled dataset (e.g. BUC \cite{lin2019bottom}, SSL \cite{lin2020unsupervised}, DBC \cite{ding2019towards}, HCT \cite{zeng2020hierarchical}, MMCL \cite{wang2020unsupervised}); 3) the unsupervised transfer learning and domain adaptation approaches (e.g. UMDL \cite{peng2016unsupervised}, CAMEL \cite{yu2017cross}, PUL \cite{fan2018unsupervised}, PTGAN \cite{wei2018person}, SPGAN+LMP \cite{deng2018image}, MMFA \cite{lin2018multi}, TJ-AIDL \cite{wang2018transferable}, HHL \cite{zhong2018generalizing}, ECN \cite{zhong2019invariance}, MAR \cite{yu2019unsupervised}, PAUL \cite{yang2019patch}, SSG \cite{fu2019self}, CR-GAN \cite{chen2019instance}, CASCL \cite{wu2019unsupervised}, PDA-Net \cite{li2019cross}, MMCL \cite{wang2020unsupervised}, ADTC \cite{wuattention2020}, and D-MMD \cite{mekhazni2020unsupervised}). The comparison results on Market-1501 and DukeMTMC-reID are presented in Table \ref{tab:table2}, and the comparison on MSMT17 is listed in Table \ref{tab:table3}.

From Table \ref{tab:table2}, we observe that our MLC method consistently outperforms recent state-of-the-art methods on both Market-1501 and DukeMTMC-reID with or without source dataset. 
Without source dataset, the hand-crafted feature based methods show the worst performance since the representation ability of designed features is limited.
These deep learning methods (from BUC to MMCL) with pseudo labels significantly outperform the hand-crafted feature based methods which indicates that pseudo labels and deep networks are effective. Our MLC achieves the best results with mAP 66.2\% and 52.3\% on Market-1501 and DukeMTMC-reID. Compared with pure clustering based methods like SSL, our MLC leverages the ML module to learn better initial representations to guide subsequent clustering. Compared with the MMCL method, our MLC uses another SC module to generate new dataset and refine pseudo labels which largely enhances the Re-ID models through joint training.

\pxj{We further compare our method with unsupervised transfer learning and domain adaptation methods. Our MLC not only surpasses those fully unsupervised methods  but is also better than these unsupervised transfer learning and domain adaptation methods. Specifically, under the transfer learning setting, our method achieves the best performance on both Market1501 and DukeMTMC-reID. For example, our MLC obtains mAP 65.9\% and 55.0\% on Market1501 and DukeMTMC-reID, respectively. An interesting observation is that the results of our MLC are similar whether there exists annotated source dataset or not. This indicates that our method is suitable for unsupervised person Re-ID learning.}

\pxj{We also conduct experiments on MSMT17, and the results are shown in Table \ref{tab:table3}. Compared with the other two datasets, MSMT17 is a larger and more challenging dataset because of more complex lighting and scene variations. A limited number of researches have reported performance on MSMT17, including unsupervised learning and transfer learning with domain adaptation methods, such as MMCL \cite{wang2020unsupervised}, PTGAN \cite{wei2018person}, ECN \cite{zhong2019invariance}, and SSG\cite{fu2019self}. As shown in Table \ref{tab:table3}, our approach outperforms existing methods by large margins under both unsupervised and transfer learning methods. We can see that our MLC without MN module obtain the best performance on unsupervised learning method. It achieves mAP 12.0\% and outperforms our MLC and MMCL by 0.7\% and 1.5\% improvements, respectively. Under transfer learning setting, our MLC obtains mAP 15.0\% and 16.7\% with the source datasets of Market1501 and DukeMTMC-reID. Interestingly, the MLC without MN module achieves better results with mAP 16.2\% and 18.0\%, respectively. The slightly improve of MN module on MSMT17 may be explained by that the person images of MSMT17 are more variant than the other two datasets in posture and scene. Overall, our MLC improves the state-of-the-arts (\textit{i.e.} MMCL) by 1.8\% and 2.8\% in mAP and Rank-1 when using Duke as source dataset.}

\begin{table}[htp]
\centering
\caption{Performance ($\%$) comparison of our framework with state-of-the-art methods on MSMT17.}\label{tab:table3}
\footnotesize
\begin{tabular}{l|c|c|c|c|c}
\toprule
 \hline
 \multicolumn{1}{c|}{\multirow{2}*{Method}} &  \multicolumn{1}{c|}{\multirow{2}*{source}} & \multicolumn{4}{c}{MSMT17}\\
 \cline{3-6}
  \multicolumn{1}{c|}{} & \multicolumn{1}{c|}{} & Rank-1 & Rank-5 & Rank-10 & mAP \\
 \hline
 MMCL \cite{wang2020unsupervised} & None & 35.4 & 44.8 & 49.8 & 11.2 \\
 MLC w/o MN & None & \textbf{39.2} & \textbf{49.4} & \textbf{53.9} & \textbf{12.7} \\
 MLC & None & 37.1 & 47.5 & 52.4 & 12.0 \\
  \hline
 PTGAN \cite{wei2018person} & Market & 10.2 & - & 24.4 & 2.9 \\
 ECN \cite{zhong2019invariance} & Market & 25.3 & 36.3 & 42.1 & 8.5 \\
 SSG\cite{fu2019self} & Market & 31.6 & - & 49.6 & 13.2 \\
 MMCL \cite{wang2020unsupervised} & Market & 40.8 & 51.8 & 56.7 & 15.1 \\
 MLC w/o MN & Market & 42.4 & 53.0 & 57.9 & 16.2 \\
 MLC & Market & \textbf{43.9} & \textbf{55.3} & \textbf{60.4} & \textbf{16.5} \\
 \hline
  PTGAN \cite{wei2018person} & Duke & 11.8 & - & 27.4 & 3.3 \\
 ECN \cite{zhong2019invariance} & Duke & 30.2 & 41.5 & 46.8 & 10.2 \\
 SSG\cite{fu2019self} & Duke & 32.2 & - & 51.2 & 13.3 \\
 MMCL \cite{wang2020unsupervised} & Duke & 43.6 & 54.3 & 58.9 & 16.2 \\
 MLC w/o MN & Duke & 45.0 & 55.9 & 60.8 & 16.7 \\
 MLC  & Duke & \textbf{46.4} & \textbf{57.9} & \textbf{62.7} & \textbf{18.0} \\
 \hline
\bottomrule
\end{tabular}
\end{table}

\subsection{Ablation study on Market1501 and DukeMTMC-reID}

\begin{table*}[htp]
\centering
\caption{The performance ($\%$) comparison of proposed individual components of MLC on Market-1501 dataset and DukeMTMC-reID.}\label{tab:table4}
\begin{tabular}{ccc|c|c|c|c|c|c|c|c}
\toprule
 \hline
 \multicolumn{3}{c|}{Module} & \multicolumn{4}{c|}{ Market-1501} & \multicolumn{4}{c}{DukeMTMC-reID} \\
 \hline
  MN & ML & SC & Rank-1 & Rank-5 & Rank-10 & mAP & Rank-1 & Rank-5 & Rank-10 & mAP \\
 \hline
 $\surd$ & $\surd$ &k-means(k=500) & 80.5 & 89.9 & 92.4 & 49.1 & 65.0 & 76.4 & 80.9 & 41.6 \\
  $\surd$ & $\surd$ &k-means(k=750) & 81.8 & 90.3 & 92.8 & 49.3 & 64.5 & 76.4 & 80.5 & 41.3 \\
  $\surd$ & $\surd$ &k-means(k=1000) & 81.5 & 89.8 & 92.4 & 48.5 & 65.2 & 76.0 & 80.8 & 42.1  \\
  $\times$ & $\surd$ &$\times$ & 79.8 & 88.4 & 91.6 & 44.7 & 65.6 & 75.9 & 80.1 & 39.6  \\
  $\surd$ & $\surd$ & $\times$& 79.7 & 89.3 & 92.3 & 44.1 & 66.1 & 76.8 & 80.1 & 41.5  \\
  $\times$ & $\surd$  & DBSCAN & 85.2 & 92.2 & 94.6 & 62.6 & 71.9 & 81.2 & 84.4 & 50.0  \\
 $\surd$ & $\surd$  & DBSCAN & \textbf{86.7} & \textbf{93.5} & \textbf{95.6} & \textbf{66.2} & \textbf{73.6} & \textbf{82.3} & \textbf{85.5} & \textbf{52.3}  \\
 \hline
\bottomrule
\end{tabular}
\end{table*}

\textbf{Evaluation of individual modules in MLC}.
In order to verify the effectiveness of each module in MLC, we conduct the experiments by evaluating the performance contribution of different modules on the Market-1501 and DukeMTMC-reID dataset. It is summarized in Table \ref{tab:table4}. We utilize the k-means instead of the DBSCAN as the clustering method in our SC module for comparison. And the number of pseudo identities of k-means clustering are set as 500, 750, 1000. Especially, the cross (or tick) in MN, ML, SC column means the result of MLC without (or with) the corresponding modules, respectively.

Seen in Table \ref{tab:table4}, when using the k-means instead of the DBSCAN method, the best mAP is 49.3\% on Market-1501 and 42.1\% on DukeMTMC-reID by $k=750$ and $k=1000$, respectively. Seen from the results, the SC module could improve MLC's performance obviously on Market-1501 while slightly on DukeMTMC-reID. This shows that even an elementary clustering algorithm could push the SC module, which improves the network to learn more discriminative image features. It can be evidenced that using the SC module could be beneficial to all modules in our MLC. Compared with MLC without the SC module, our MLC is found to significantly boost 22.1\% and 10.8\% mAP on Market-1501 and DukeMTMC-reID. Especially, it is shown that MLC can achieve the best performance with all modules on both datasets. This result shows, on the one hand, the SC module can remove the noisy samples when joint learning with the ML module. On the other hand, it also can help our MLC to generate more accurate pseudo labels for unsupervised learning.

\begin{figure*}
\center
\includegraphics[width=0.95\textwidth]{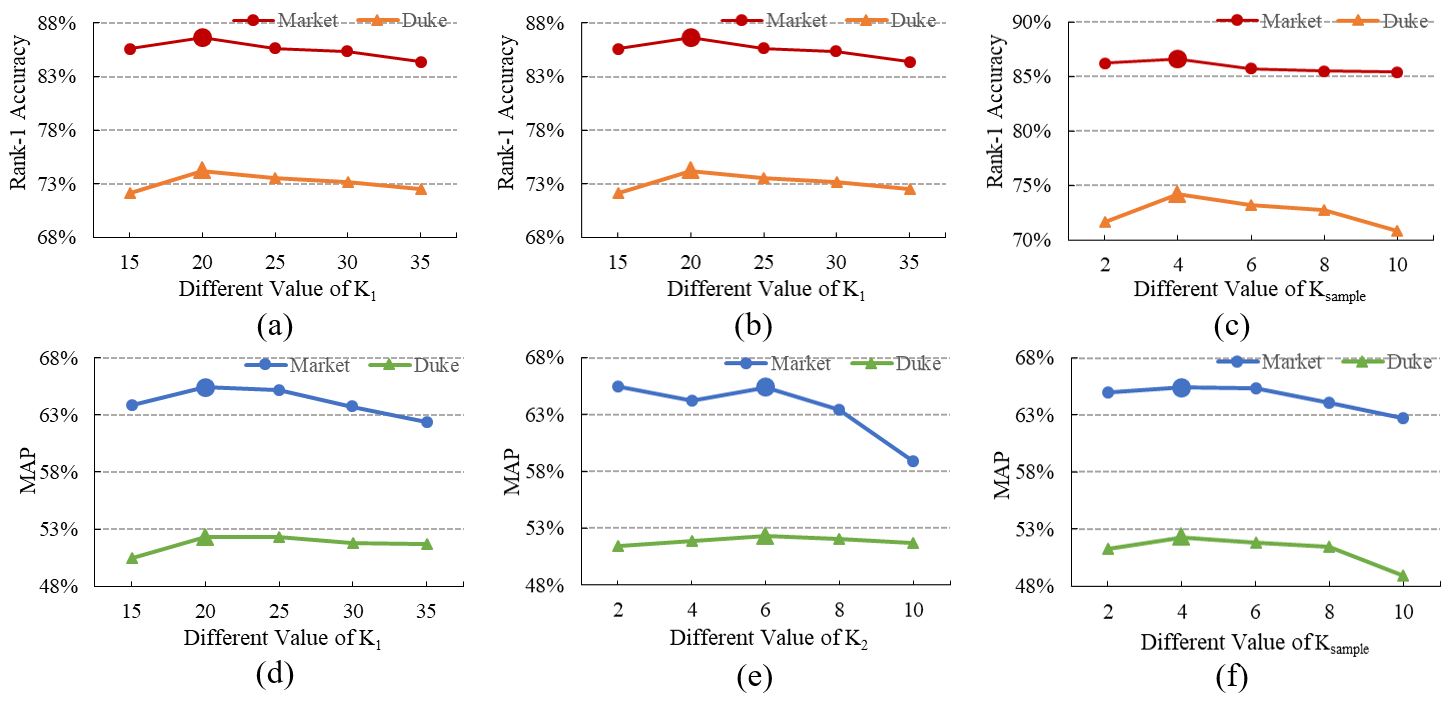}
\caption{Evaluation of the different numbers of $K_{1}$, $K_{2}$ and $K_{sample}$ on Market-1501 and DukeMTMC-reID dataset. (a), (c) and (e): rank-1 accuracy. (b), (d) and (f): mAP.}
\label{fig:K}
\end{figure*}

\begin{figure}
\center
\includegraphics[width=0.45\textwidth]{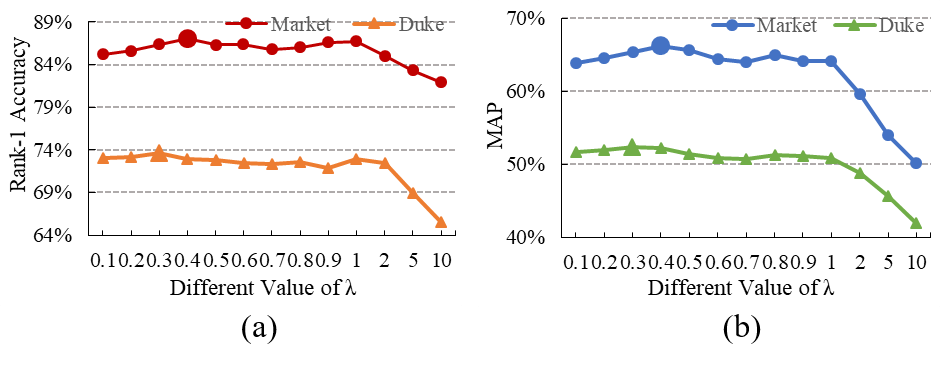}
\caption{Evaluation of the different numbers of $\lambda$ on Market-1501 and DukeMTMC-reID dataset. (a) : rank-1 accuracy. (b): mAP.}
\label{fig:lamda}
\end{figure}

\begin{figure*}
\center
\includegraphics[width=0.75\textwidth]{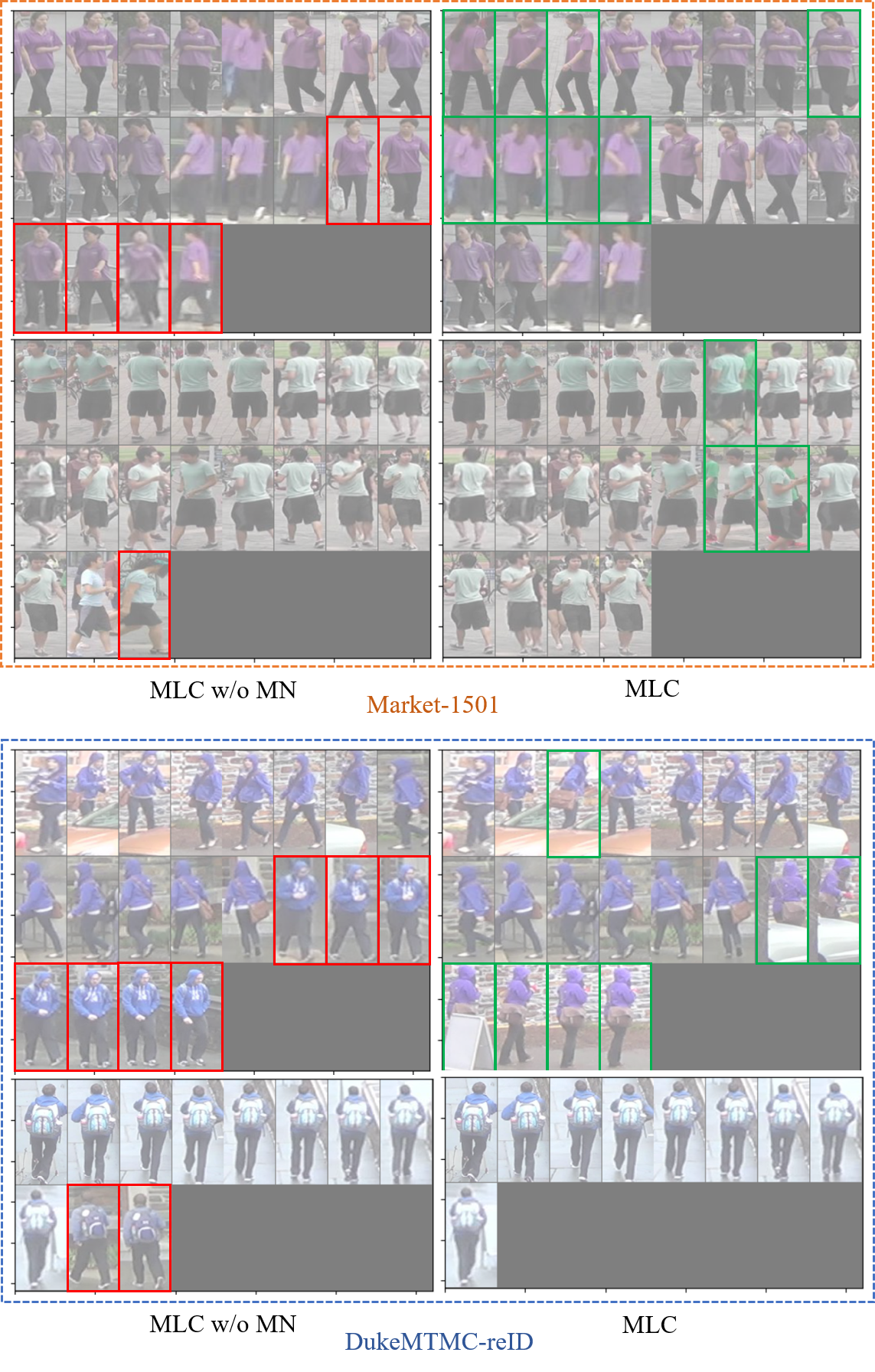}
\caption{Visualization of the clustering performances by our MLC with and without MN module are both on the Market-1501 and DukeMTMC-reID dataset. The clustering samples with the top 20 nearest neighbors are obtained by the MLC w/o MN and MLC method, which are shown in the left column and the right column, respectively. The image with a red bounding box means a noisy example in that clustering group, and the rest are correct examples. More specifically, the image with a green bounding box means a positive sample only found by MLC but not the other.}
\label{fig:visual}
\end{figure*}

\textbf{Evaluation of the parameter $K_{1}$, $K_{2}$ and $K_{sample}$}.
In the clustering phase, the $K_{1}$ and $K_{2}$ is denoted the parameter in evaluating of the Jaccard distance \cite{zhong2017re}, the former is considered as the contextual knowledge to re-calculate the distance between the probe and gallery, the latter is the number of candidates in the top-k samples of the ranking list about the probe. And the parameter $K_{sample}$ is the minimum sample number of groups in the DBSCAN method. To investigate the properties of our MLC with these parameters, we fix the hyperparameter $\lambda$ as 0.3. And the default setting of $K_{1}$, $K_{2}$ and $K_{sample}$ are 20, 6 and 4, respectively. We evaluate it by increasing $K_{1}$ from 15 to 35, and present the results in Figure \ref{fig:K} (a) and (d). It can be seen that the performance first increases with the growth of $K_{1}$, and then begins a slow decline after $K_{1}$ set to 20. Larger K1 means it is more likely to include false matches in the k-reciprocal set, resulting in a decline in performance. The impact of the size of $K_{2}$ is shown in Figure \ref{fig:K} (b) and (e), it is varied from 2 to 10. We can see that when we set $K_{2}=6$, we get the best performance. Notice that, assigning a too large value to $K_{2}$ also reduces the performance. At last, we evaluate $K_{sample}$ from 2 to 10. Our results are reported in Figure \ref{fig:K} (c) and (f), the performance grows as $K_{sample}$ increases in a reasonable range. Value of $K_{sample}$ greater than 4 will reduce the performance. If the parameter of $K_{sample}$ is set too large, it will affect the performance of DBSCAN clustering.

\textbf{Evaluation of the hyperparameter $\lambda$}.
We evaluate how $\lambda$ (the relative importance of multi-label classification loss) affects our model learning. To investigate the properties of our MLC framework, we evaluate different $\lambda$ from 0.1 to 10. The results of rank-1 accuracy and mAP on Market-1501 and DukeMTMC-reID dataset are shown in Figure \ref{fig:lamda} (a) and (b). We observe that increasing $\lambda$ boosts performance in the beginning, after $\lambda=0.4$ on Market-1501 and $\lambda=0.3$ on DukeMTMC-reID gradually degrades. Too larger $\lambda$ leads to fast degradation, which indicates the pseudo label classification loss is important for joint training our model. As the mater of fact, the result of this experiment is evidenced that the SC joints in training with the ML is a simple and effective strategy for unsupervised person Re-ID tasks.

\textbf{Qualitative analysis of clustering visualization}
To better investigate the effectiveness of our MLC, we visualize the clustering results in Figure \ref{fig:visual}. We illustrate the clustering performance of our MLC with and without the MN module in the right and left clomun, respectively. We also shows the results of MLC (w or w/o the MN module) on  Market-1501 and DukeMTMC-reID dataset. The cluster image shows that the MN module could help not only purifying negative samples (images with red bounding box), but also supplementing hard positive samples (images with green bounding box) in each group.This indicates that our MN module can improve the clustering quality by mining the multi-scale features for better similarity measurement.

\section{Conclusion}
\pxj{In this paper, we proposed a novel multi-label learning guided self-paced clustering (MLC) framework for unsupervised Person Re-identification (Re-ID). It mainly contains three modules: the multi-scale network which obtains global and local person representations, the multi-label learning module which trains the network with memory bank and multi-label classification loss, and the self-paced clustering module which removes noisy samples and assigns pseudo labels for training. Extensive experiments on three challenging large-scale datasets demonstrated the effectiveness of all the modules. Our MLC framework finally achieves state-of-the-art performance on these datasets.} 

{\small
\bibliographystyle{ieee}
\bibliography{ref}
}

\end{document}